\documentclass[11pt, conference]{IEEEtran}
\IEEEoverridecommandlockouts
\usepackage[a4paper,left=1.6cm,right=1.6cm,top=1.9cm,bottom=4.3cm]{geometry}

\usepackage{amsmath} 
\usepackage{amssymb}  
\usepackage{mathtools}
\usepackage{upgreek}
\usepackage{color,soul}
\usepackage{lipsum}
\usepackage{float}
\usepackage{balance}
\usepackage{url}
\usepackage{xr}
\usepackage{algpseudocode}
\usepackage{xcolor}
\usepackage{breqn}

\usepackage[most]{tcolorbox}
\usepackage{booktabs}
\usepackage{cite}
\usepackage{graphicx}
\usepackage[compatibility=false]{caption}
\usepackage{graphics} 
\usepackage{subcaption} 
\usepackage{epsfig} 
\usepackage{gensymb}
\usepackage[ruled,longend,linesnumbered]{algorithm2e}
\algnewcommand{\algorithmicand}{\textbf{ and }}
\algnewcommand{\algorithmicor}{\textbf{ or }}
\algnewcommand{\algorithmicnot}{\textbf{ not }}
\algnewcommand{\OR}{\algorithmicor}
\algnewcommand{\AND}{\algorithmicand}
\algnewcommand{\NOT}{\algorithmicnot}

\usepackage{amssymb}
\usepackage{amsthm}
\usepackage{placeins}

\title{LLM-based Multi-Criteria Deep Reinforcement Learning
 for Dynamic Efficient 5G UPF Selection}

\author{\author{ Abdelrahman Soliman, Ahmed Refaey Hussein, Lei Lei \\University of Guelph, Guelph, Ontario, Canada}
	
}

\begin{document}

\maketitle

\begin{abstract}
User plane function (UPF) selection in 5G networks is an active area of research, with various limited approaches proposed to enhance network efficiency and manage resources effectively. In this paper, we address this gap in the related literature by proposing a Deep Reinforcement Learning (DRL) UPF selection method that utilizes a dynamic, multi-objective reward function and validating it on a real-world 5G emulation platform. The proposed method is based on the large language models' (LLMs) translation of network operator intents and the network data analytics function (NWDAF) abnormality alerts. The experimental results validated in the real-world experimental setup demonstrate that our dynamic DRL solution achieves an effective balance between latency, power consumption, and traffic distribution, outperforming the heuristic baselines and fixed DRL by 16.89\%. These findings highlight the real-world applicability of our approach for dynamic, intent-driven UPF selection.
\end{abstract}

\begin{IEEEkeywords}
	\makeatletter
	\def\RemoveSpaces#1{\zap@space#1 \@empty}
	\makeatother 
5G,
Core Network,
Large Language Models,
Deep Reinforcement Learning,
Intelligent Networks,
Edge Computing,
\end{IEEEkeywords}

\section{Introduction}
\label{sec:intro}
The current rapid increase in demands and requirements in Fifth Generation (5G) and next-generation networks has led mobile network operators to focus on sustainability and energy efficiency. Current radio access networks (RAN) consume around 70-80\% \cite{doorgakant2025end} of total energy consumption, while the core network consumes 20-30\%. These numbers reflect the need to optimize energy usage and implement intelligent, cost-effective solutions.

One critical innovation in the 5G architecture is the principle of control and user plane separation, where we decouple control plane functions, such as the Session Management Function (SMF), from the user plane, represented by the UPF. This separation allows the UPF to be deployed and scaled independently of the control plane. Consequently, operators can distribute UPF instances across a wide geographical area, placing them in centralized cloud data centers and at the network edge in Multi-access Edge Computing (MEC) environments \cite{husain20223gpp}. Though this division offers advantages, it also presents a major orchestration challenge, which is dynamically selecting the optimal UPF for each user session from a set of heterogeneous, geographically distributed instances. This decision must balance multiple competing objectives, such as minimizing power consumption and meeting stringent latency and traffic requirements.

In the era of artificial intelligence (AI), several research papers have highlighted different methods in the process of UPF selection. Initial research focused on algorithms for dynamic selection \cite{alevizaki2021dynamic} and optimal reconfiguration \cite{leyva2020dynamic} to adapt to changing network conditions. More recent studies have specifically adopted Deep Reinforcement Learning (DRL) as a promising tool for this complex optimization task. For instance,  \cite{yokobori2019user} demonstrated the use of DRL for user plane selection in core networks. \cite{bellin2024power} further advanced this by applying DRL to select edge UPFs, with a specific focus on power consumption awareness. Concurrently, work by \cite{barrachina2024demand} has explored on-demand UPF selection and relocation to support 5G/6G edge verticals.

The current research presents promising solutions, but it also contains multiple gaps that our study aims to address. The first problem with current DRL models stems from their single-objective design, which restricts them to optimizing power metrics \cite{bellin2024power} but fails to address the multiple trade-offs among power consumption, latency, and traffic management. The novelty of this work lies in the operational integration of three components within the UPF-selection loop: intent and NWDAF-alert interpretation using an LLM, conversion of these high-level inputs into an explicit multi-objective preference vector, and a PPO-based UPF-selection policy conditioned on that vector and on live network metrics. The framework is further evaluated in a 5G-core emulation rather than solely in a simulation environment.

The contributions of this paper can be summarized as follows:
\begin{itemize}
	  \item We propose a multi-criteria DRL for the efficient selection between multiple edge and cloud UPFs while considering the trade-off among power, latency, and traffic. 
    \item To further increase the adaptability of the method, we connected the DRL with LLMs to support dynamic selection through anomaly alerts and network operator intents. 
    \item We provide comparative experiments and analyzes of our approach using a modified 5G core network emulator and UE-RAN simulator.
  
\end{itemize}  

The remainder of this paper is organized as follows: Section II reviews related work, Sections III and IV detail the system model and proposed approach, Section V discusses experimental results, and Section VI concludes.

\section{Related Work}
Early UPF placement strategies relied on centralized optimization, such as the Integer Linear Programming (ILP) and scheduling techniques proposed by \cite{leyva2020dynamic} to minimize costs and latency. However, these scheduled approaches lack real-time adaptability. To enable dynamic, decentralized selection, \cite{alevizaki2021dynamic} applied Evolutionary Game Theory (EGT), allowing User Equipment (UEs) to independently select UPFs to minimize service delay and adapt to traffic loads without centralized control.

The application of AI, particularly DRL, has been applied to dynamic UPF selection because it can adapt to changing network conditions. Work in \cite{yokobori2019user} improved DRL learning in large networks using Prioritized Experience Replay (PER) and Hindsight Experience Replay (HER) to address sparse rewards. Building on this, work in \cite{bellin2024power} proposed a DRL-based UPF orchestrator that minimizes power consumption using real-time edge and cloud metrics while satisfying QoS requirements.

As shown in the comparison table \ref{tab:comparison}, while these works have made promising contributions, they often focus on a single objective (latency or power) or are primarily validated through simulation. We extend these foundations by presenting an adaptive, intent-driven LLM multi-criteria DRL framework validated through real-time 5G core emulation.

\begin{table}[h]
\renewcommand{\arraystretch}{1.4} 
\setlength{\tabcolsep}{2.5pt} 
\centering
\caption{Comparison of UPF Selection Methods in Related Work}
\label{tab:comparison}
\begin{tabular}{|p{0.7cm}|p{1.5cm}|c|p{0.8cm}|c|p{1.9cm}|}
\hline
\textbf{Ref} & \textbf{Method} & \textbf{Objective} & \textbf{Intent} & \textbf{Adaptive} & \textbf{Validation} \\
\hline
\cite{leyva2020dynamic} & ILP / Heuristic & Multi & $\times$ & $\times$ & Simulation \\
\hline
\cite{barrachina2024demand} & System Architecture (PoC) & N/A & $\times$ & $\times$ & Emulation (PoC) \\
\hline
\cite{yokobori2019user} & DRL & Single & $\times$ & \checkmark & Simulator (physical UP) \\
\hline
\cite{bellin2024power} & DRL & Single & $\times$ & \checkmark & Simulation (data-driven) \\
\hline
\cite{alevizaki2021dynamic} & Game Theory & Single & $\times$ & \checkmark & Simulation \\
\hline

\textbf{Ours} & \textbf{LLM-based Multi-criteria DRL} & \textbf{Multi} & \textbf{\checkmark} & \textbf{\checkmark} & \textbf{Real 5G Core Emulation} \\
\hline
\end{tabular}
\end{table}

\section{System Model}
\begin{figure}[h] 
    \centering
    \includegraphics[width=\linewidth]{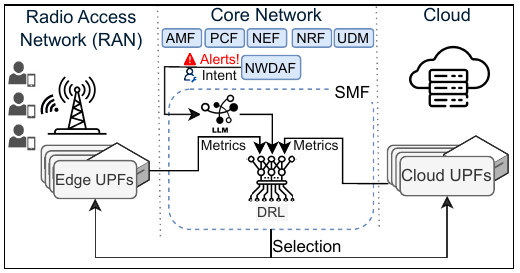}
    \caption{System model showing the UPFs selection through DRL and enhanced by LLMs.}

    \label{fig:arch}
\end{figure}

As shown in Figure \ref{fig:arch}, the UPF selection process occurs in the SMF, where it receives a protocol data unit (PDU) session establishment procedure that connects the UE to the data network through UPFs \cite{3gpp}. In this system, several edge and cloud UPF candidates exist, and the main goal is to select the best UPF that either balances power, latency, and traffic or prioritizes specific metrics based on the LLM's translation of network operator intents or the network data analytics function (NWDAF) abnormality alerts. 

The edge UPFs are usually placed close to the user side, such as in MEC. They are capable of offering ultra-low latency and high bandwidth by processing data close to the end-user \cite{goshi2024joint}. This local data breakout is essential for use cases like augmented reality (AR) and autonomous vehicle communication, as it significantly reduces the traffic sent over the backhaul network to the central core.

For the cloud UPFs, the placement and function serve different purposes. They are centralized within regional data centers and act as the main gateway to the internet for general-purpose traffic. Cloud UPFs are designed to handle a massive volume of aggregated data from a broad geographic area. While they introduce higher latency due to the geographical distance from the user, they are ideal for services that are not delay-sensitive, such as standard web browsing and non-real-time data transfers.

\section{Proposed approach}
\begin{figure}[h] 
    \centering
    \includegraphics[width=1\linewidth]{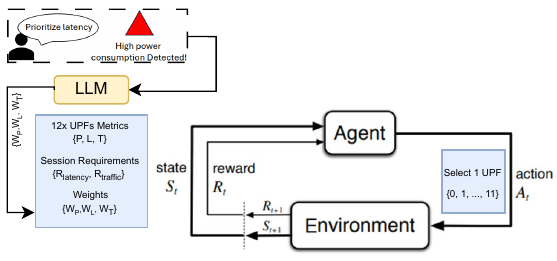}
    \caption{Proposed DRL approach overview, presenting the state and action space.}

    \label{fig:drl}
\end{figure}

In this section, we will discuss the two main components of our proposed method, which are the DRL and LLMs. As shown in Figure \ref{fig:drl}, the DRL will be responsible for UPF selection, while the LLMs will translate the network operator's intents or abnormality alerts.
\subsection{Large Language Model}
We used a large language model for advanced intent translation and adaptive alert resolution. These intents help network operators prioritize metrics during UPF selection, while LLMs leverage analytics and alerts from NWDAF for the selection process. The LLM outputs the weights \{$w_p$, $w_l$, $w_t$\}, which correspond to power, latency, and traffic, respectively. These weights are then forwarded to the state space, together with relevant metrics and requirements for inference. With the few-shot system prompt shown below, we leveraged Meta Llama 3.1, a small-scale LLM model with 8 billion parameters. Before being added to the DRL state, the LLM output is validated to ensure it contains three finite, non-negative weights. The weights are normalized so that their sum equals one. If the output is invalid or malformed, it is returned to the LLM for regeneration.

\begin{tcolorbox}[colback=gray!14!white, colframe=gray, title=\textbf{LLM System Prompt}]
Given user intent/NWDAF Alert: "\{Input\}."

Output three weights (w1, w2, w3) 

for [power, latency, traffic]
that sum to 1.0 and reflect the input goal.

Examples:

- "High power alert!" → [0.7, 0.15, 0.15]

- "Prioritize latency" → [0.15, 0.7, 0.15]

- "Balance all" → [0.33, 0.34, 0.33]
\end{tcolorbox}

\subsection{Deep Reinforcement Learning}
This section will detail the key components of the DRL solution, as shown in Figure \ref{fig:drl}.

\subsubsection{Environment Design}
The UPF selection problem is formulated as a Markov Decision Process (MDP) using the Gymnasium framework. The environment models a 5G network with a heterogeneous UPFs infrastructure consisting of edge and cloud UPFs. At each time step $t$, the agent observes the current state of all available UPFs and session requirements, selects a UPF, and receives a reward signal based on the selected UPF's performance metrics.

\subsubsection{Action and State Space}
The state space $\mathcal{S}$ as shown in Eq (1) is a continuous, 41-dimensional vector that encodes both UPF metrics and session requirements. Specifically, for each of the 12 UPFs, three normalized metrics are included: power consumption $P_i$, traffic load $T_i$, and latency $L_i$. Additionally, the state includes two session requirement parameters: required latency $L_{req}$ and required traffic threshold $T_{req}$. Finally, there are three preference weights, denoted as $(w_p, w_l, w_t)$, which correspond to the weights for power, latency, and traffic, respectively. For example, when the weights are (0.7,0.15,0.15), higher priority is given to power efficiency. It is important to note that all values are normalized to $[0, 1]$ using min-max normalization before being concatenated into the state vector:

\begin{equation}
\begin{aligned}
s_t = [&(P_1, T_1, L_1), \ldots, (P_{12}, T_{12}, L_{12}),\\
      &(L_{req}, T_{req}), (w_p,w_l,w_t)]
\end{aligned}
\end{equation}

The action space $\mathcal{A}$ is discrete, consisting of 12 possible actions, where each action $a_t \in \{0, 1, \ldots, 11\}$ corresponds to selecting a specific UPF. Actions 0-5 represent edge UPFs, while actions 6-11 represent cloud UPFs. During inference, action masking excludes inactive or unreachable UPFs from selection. The mask is updated at each decision step, allowing the PPO agent trained for up to 12 UPFs to select only among currently available edge and cloud UPFs using the current normalized network metrics, session requirements, and preference weights.

\subsubsection{Reward function}
By employing scalarization, we address the multi-criteria objectives of minimizing power consumption, latency, and traffic load using a standard, single-objective PPO agent. The reward  $r_t$ at time step $t$ is calculated as follows:
\begin{equation}
r_t =
\begin{cases}
-1, &
  \begin{aligned}[t]
  &\text{if req or capacity}\\
  &\text{no met,}
  \end{aligned}\\[4pt]
1 - (w_p\hat{P} + w_l\hat{L} + w_t\hat{T}), & \text{otherwise.}
\end{cases}
\end{equation}

where $\hat{P}$, $\hat{L}$, and $\hat{T}$ are the normalized power, latency, and traffic metrics of the selected UPF, respectively, and initially $w_p = 0.33$, $w_l = 0.34$, and $w_t = 0.33$ are the preference weights that sum to one. 

This reward structure provides a hard constraint penalty (-1) for violating session requirements or failing to meet the capacity constraint; meanwhile, it rewards UPFs for lower power consumption, traffic load, and latency when the constraints are satisfied. The fixed penalty makes constraint-violating actions less desirable than valid actions, prioritizing session latency and capacity requirements over power, latency, and traffic optimization. Larger penalties may reduce learning stability, while smaller penalties may insufficiently discourage violations.

\subsubsection{DRL Algorithm}
The agent's objective is to determine the policy that maximizes the total discounted reward.
We employ Proximal Policy Optimization (PPO) \cite{schulman2017proximal}, which falls under actor-critic on-policy algorithms. The actor, which is a policy network, selects actions based on the current state, while the critic, which is a value network, evaluates how good the action was by estimating the value of the state. PPO is chosen for its sample efficiency, stability, and ability to handle discrete action spaces effectively. The policy network $\pi_\theta(a_t | s_t)$ maps states to action probabilities using a multi-layer perceptron (MLP) with two hidden layers, each containing 256 neurons and utilizing ReLU activation functions.

\section{Experiments and Findings}
\begin{figure}[h] 
    \centering
    \includegraphics[width=0.8\linewidth]{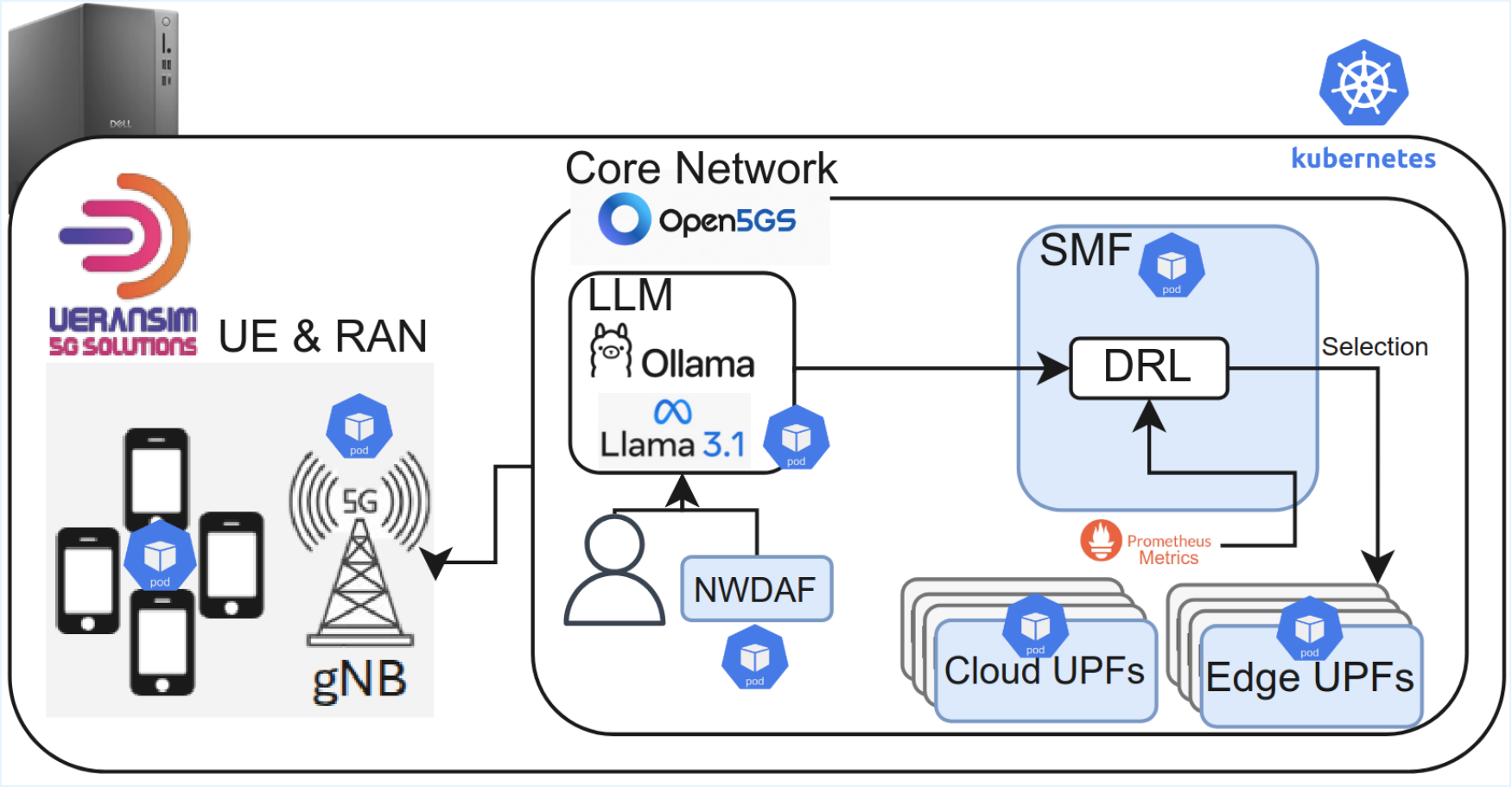}
    \caption{Experimental setup for testing the adaptive DRL UPF selection.}
    \label{fig:demo}
\end{figure}

We evaluate the proposed approach in a Kubernetes-based 5G emulation environment, shown in Figure \ref{fig:demo}. UERANSIM \cite{ueransim_2025} simulates UEs and the RAN, connected to an Open5GS core \cite{open5gs2025}. An Ollama-hosted Llama 3.1 processes intents and NWDAF alerts, while Prometheus metrics and LLM-generated weights feed the DRL agent in the SMF. Table \ref{tab:combined_env_training_params} summarizes the experimental setup and PPO parameters.

\begin{table}[t]
\centering
\caption{Summary of Experimental Environment and Training Parameters}
\label{tab:combined_env_training_params}
\begin{tabular}{lcc}
\hline
\textbf{Parameter} & \textbf{Edge UPF} & \textbf{Cloud UPF} \\ \hline
Latency & 0--2 ms & 0--4 ms \\ 
Power Consumption & 0--13 W & 0--22.5 W \\ 
Traffic Load & 0--30\% & 0--15\% \\ 
Session Capacity & 20 & 30 \\ \hline
\multicolumn{3}{l}{\textbf{Session Requirements (Uniform Distribution)}} \\ \hline
Latency Requirement & \multicolumn{2}{c}{1--20 ms} \\ 
Traffic Requirement & \multicolumn{2}{c}{1--20\%} \\ \hline
\multicolumn{3}{l}{\textbf{Global Environment Parameters}} \\ \hline
Max UPFs  & \multicolumn{2}{c}{12} \\ 
Availability Handling & \multicolumn{2}{c}{Action Masking (Inference)} \\ \hline
\multicolumn{3}{l}{\textbf{PPO Hyper-parameters}} \\ \hline
Learning Rate & \multicolumn{2}{c}{$1 \times 10^{-3}$} \\
Batch Size & \multicolumn{2}{c}{64} \\
Discount Factor ($\gamma$) & \multicolumn{2}{c}{0.99} \\
PPO Clipping Parameter ($\epsilon$) & \multicolumn{2}{c}{0.2} \\
Training Time Steps & \multicolumn{2}{c}{5 million} \\ \hline
\end{tabular}
\end{table}

\subsection{Training Results}
During training, the environment samples normalized UPF power, latency, traffic load, and session requirement values within the parameter ranges defined in Table \ref{tab:combined_env_training_params}. These ranges were selected to represent the operating envelopes observed and configured in the emulated edge and cloud UPF environment. The resulting training state has the same structure as the deployment state: per-UPF power, traffic, and latency metrics; session latency and traffic requirements; and the three intent-derived preference weights.

\begin{figure}[h] 
    \centering
    \includegraphics[width=0.9\linewidth]{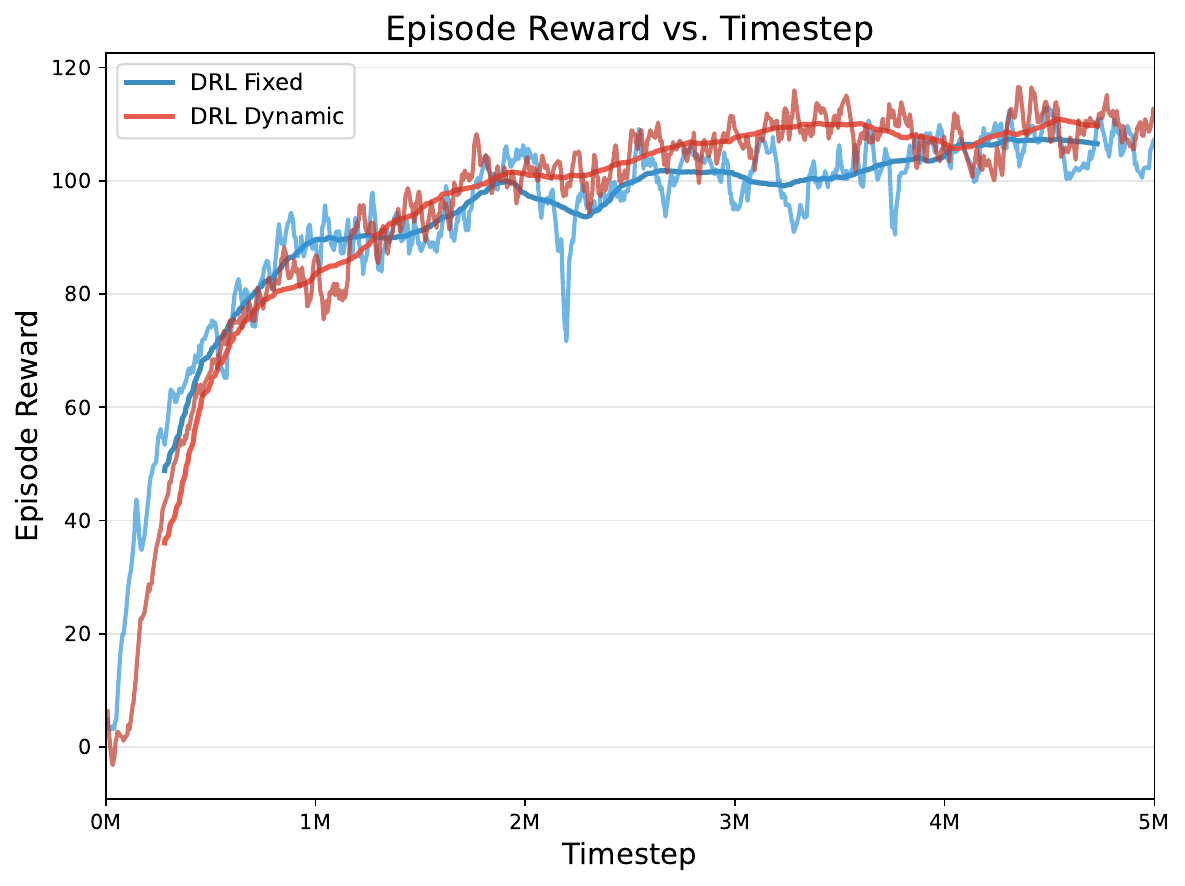}
    \caption{Episode reward vs. timestep showing the learning progress of the DRL agent over 5 million timesteps.}

    \label{fig:reward}
\end{figure}

Figure \ref{fig:reward} shows that both DRL models converge after 5 million training steps. The dynamic model achieves a higher cumulative reward, likely because training under diverse preference conditions improves exploration and policy adaptation.

\subsection{Comparison Results}
We evaluate six UPF-selection strategies: random, power-greedy, latency-greedy, traffic-greedy, fixed-DRL, and dynamic-DRL. Each method is tested in an isolated five-minute deterministic-traffic run with 10–60 UEs. Prometheus collects power, latency, and traffic metrics while enforcing capacity limits of 20 sessions for edge UPFs and 30 for cloud UPFs, along with heterogeneous session requirements. Unavailable or capacity-constrained UPFs are excluded from selection. The greedy baselines optimize a single metric, whereas fixed- and dynamic-DRL optimize the learned multi-objective reward under their respective preference weights.

\begin{figure*}[h]
\centering
\begin{subfigure}[t]{0.32\textwidth}
    \centering
    \includegraphics[width=\linewidth]{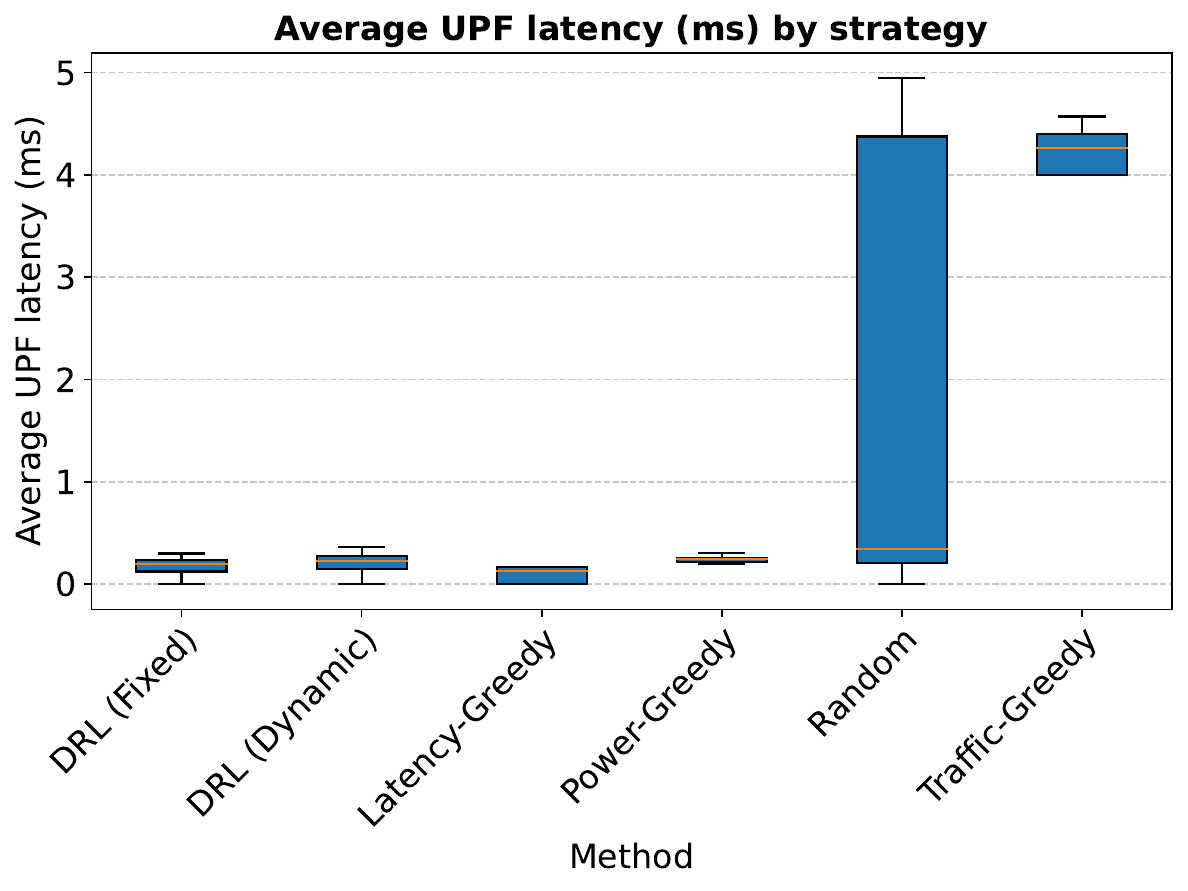}
    \caption{Latency comparison}
    \label{fig:latency_boxplot}
\end{subfigure}
\hfill
\begin{subfigure}[t]{0.32\textwidth}
    \centering
    \includegraphics[width=\linewidth]{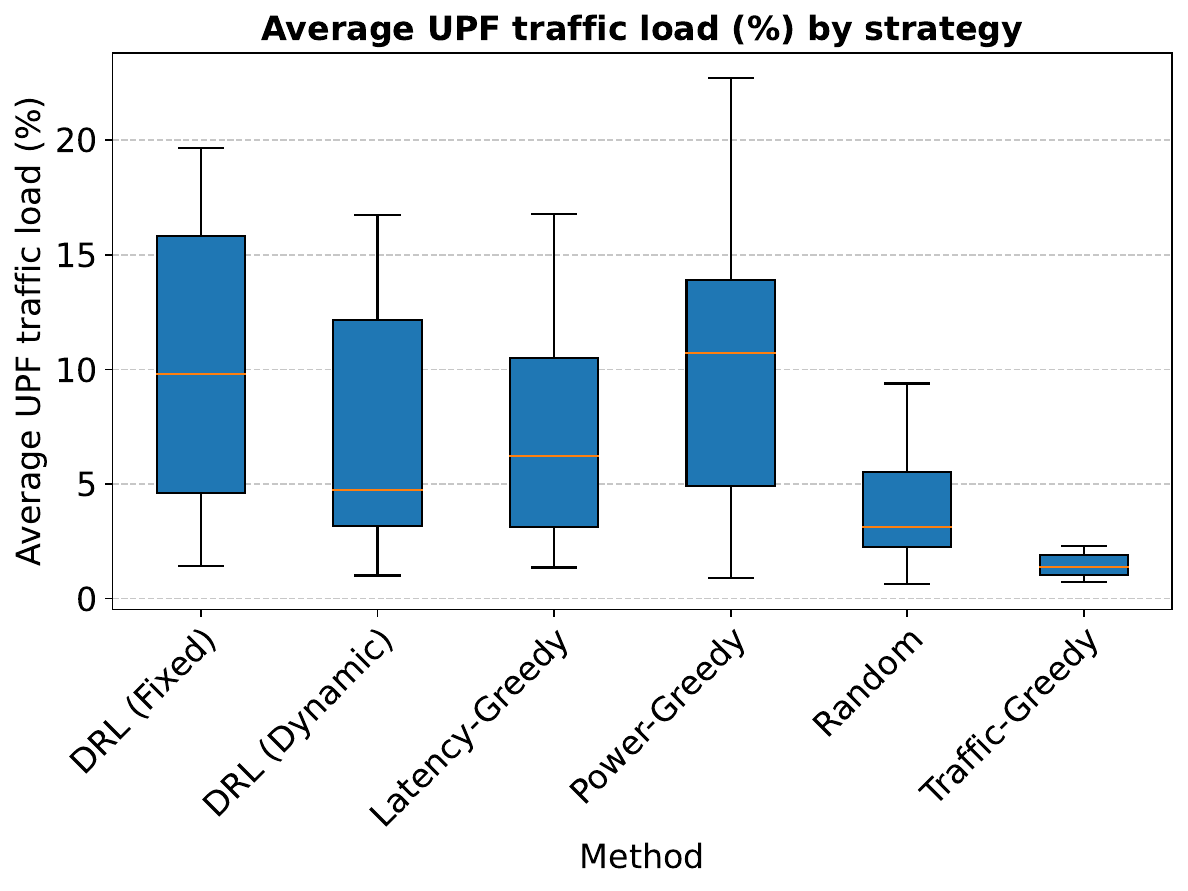}
    \caption{Traffic comparison}
    \label{fig:traffic_boxplot}
\end{subfigure}
\hfill
\begin{subfigure}[t]{0.32\textwidth}
    \centering
    \includegraphics[width=\linewidth]{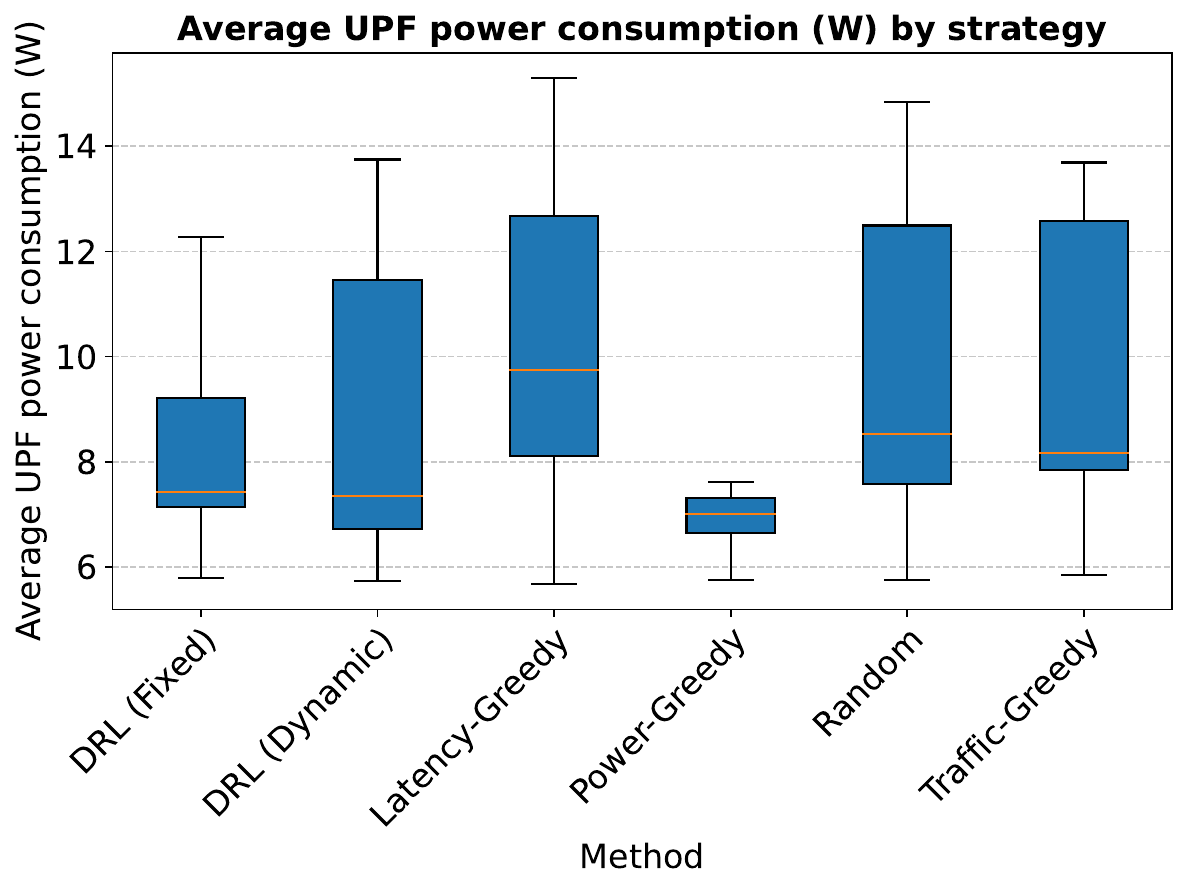}
    \caption{Power comparison}
    \label{fig:power_boxplot}
\end{subfigure}

\vspace{0.5em} 

\begin{subfigure}[t]{0.4\textwidth}
    \centering
    \includegraphics[width=\linewidth]{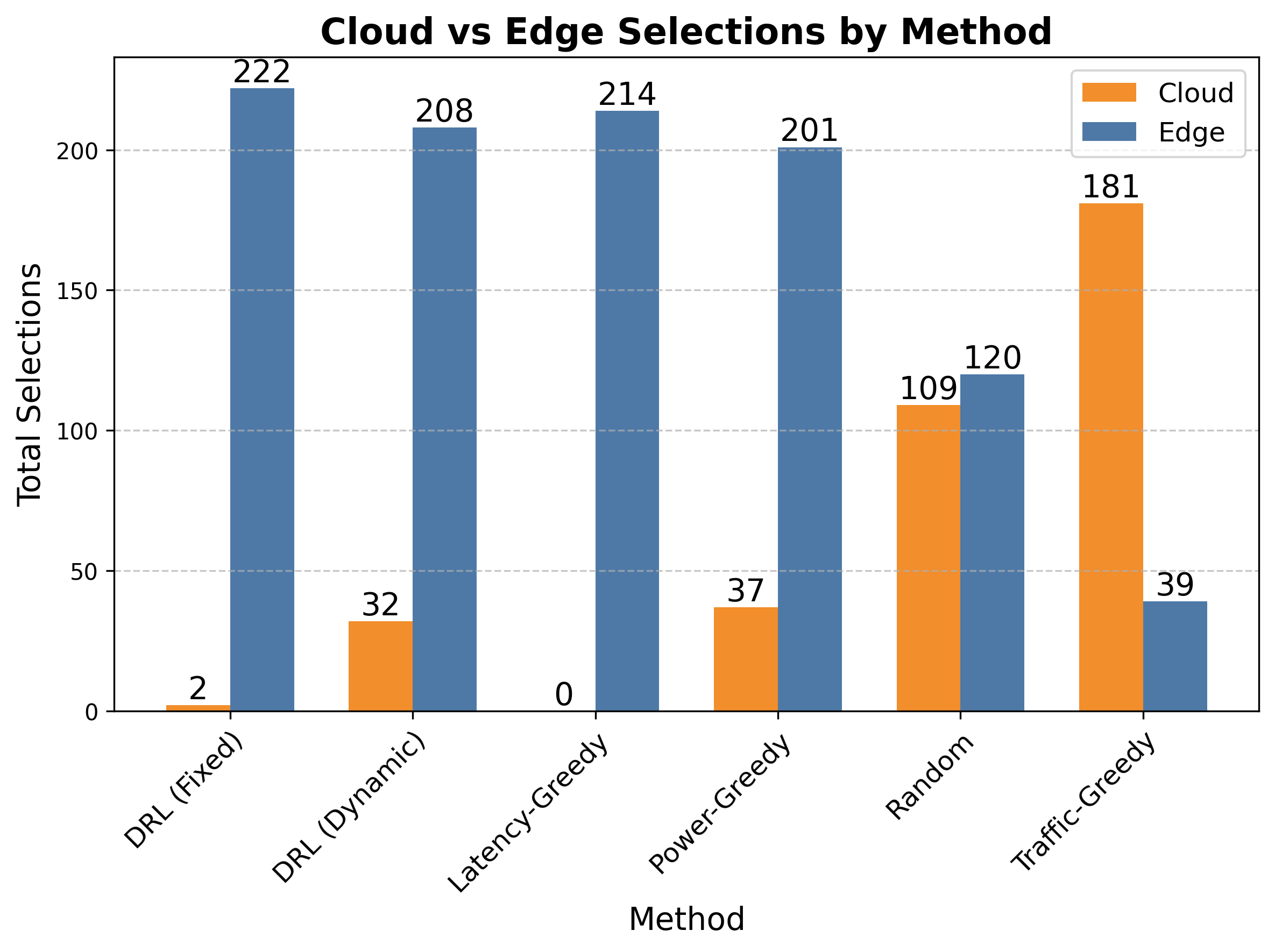}
    \caption{Selection outcomes}
    \label{fig:outcomes_bar}
\end{subfigure}
\hfill
\begin{subfigure}[t]{0.4\textwidth}
    \centering
    \includegraphics[width=\linewidth]{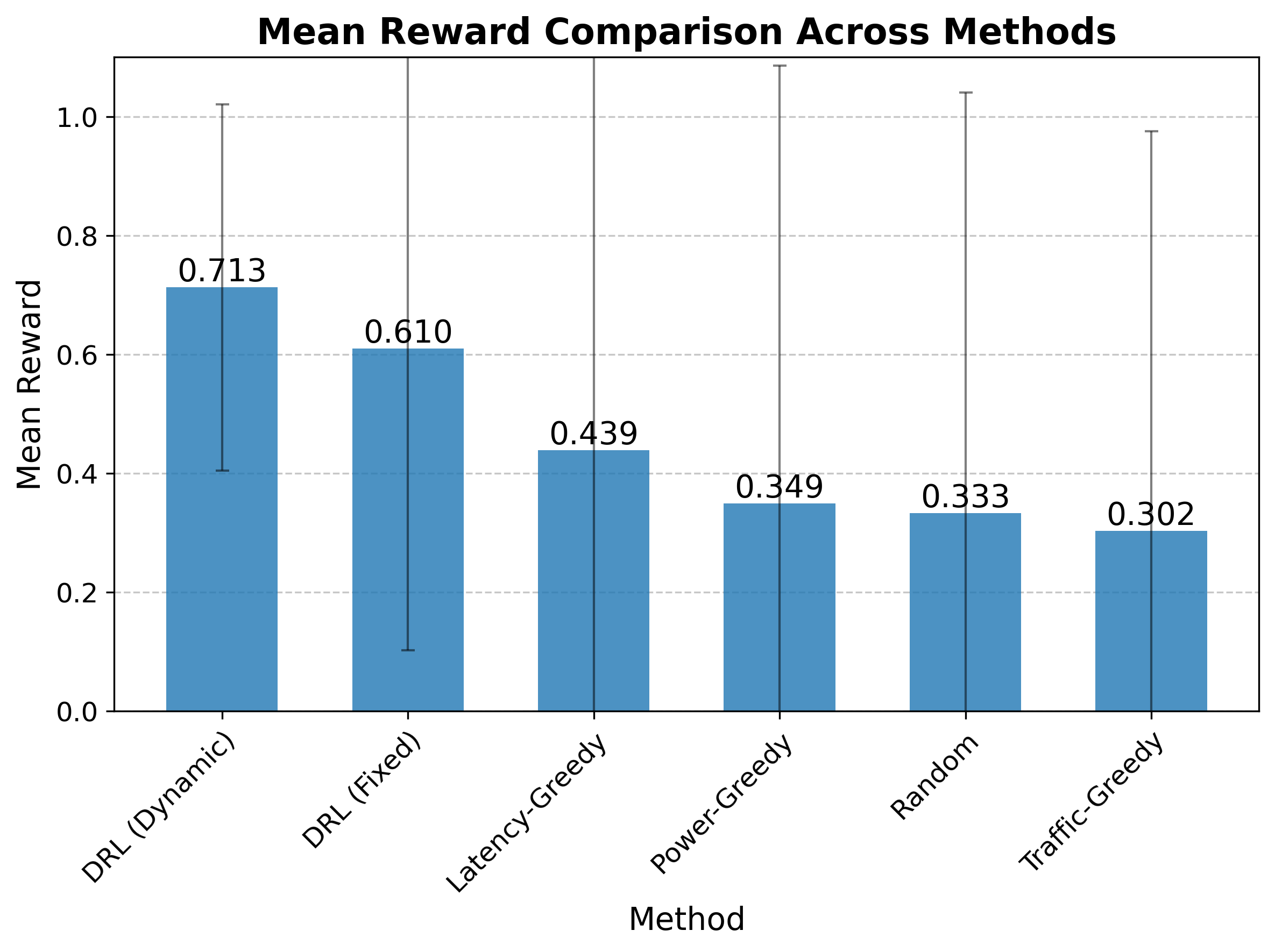} 
    \caption{Mean reward comparison}
    \label{fig:reward_mean}
\end{subfigure}

\caption{Comparative analysis of UPF selection strategies across latency, power consumption, traffic load, selection outcomes, and average reward.}
\label{fig:upf_selection_comparison}
\end{figure*}

The comparative results in Figure \ref{fig:upf_selection_comparison} highlight notable performance patterns among the UPF selection strategies. The DRL-based approaches achieved the second-lowest average UPF latency (Fig. \ref{fig:latency_boxplot}). For the traffic load analysis in Fig. \ref{fig:traffic_boxplot}, the dynamic DRL receives a high traffic load alert from NWDAF and starts selecting low-traffic cloud UPFs, which leads to a lower average traffic load. Similarly, for power consumption (Fig. \ref{fig:power_boxplot}), the power-greedy approach achieves the lowest power but incurs increased latency, whereas dynamic DRL attains comparably low power consumption without degrading the other metrics. 

The traffic-greedy method favored cloud UPFs more aggressively (Fig. \ref{fig:outcomes_bar}), resulting in higher traffic concentration and increased latency. While the random and greedy methods achieved lower averages in the metrics, their rewards (Fig. \ref{fig:reward_mean}) were very low due to the large number of session requirement violations. Meanwhile, the DRL strategy, specifically the dynamic one, demonstrated an intelligent selection between cloud and edge nodes, leading to a 16.89\% mean reward improvement compared to fixed DRL. 

\subsection{Intent/Alerts Demonstration}
\begin{figure}[H] 
    \centering
    \includegraphics[width=\linewidth]{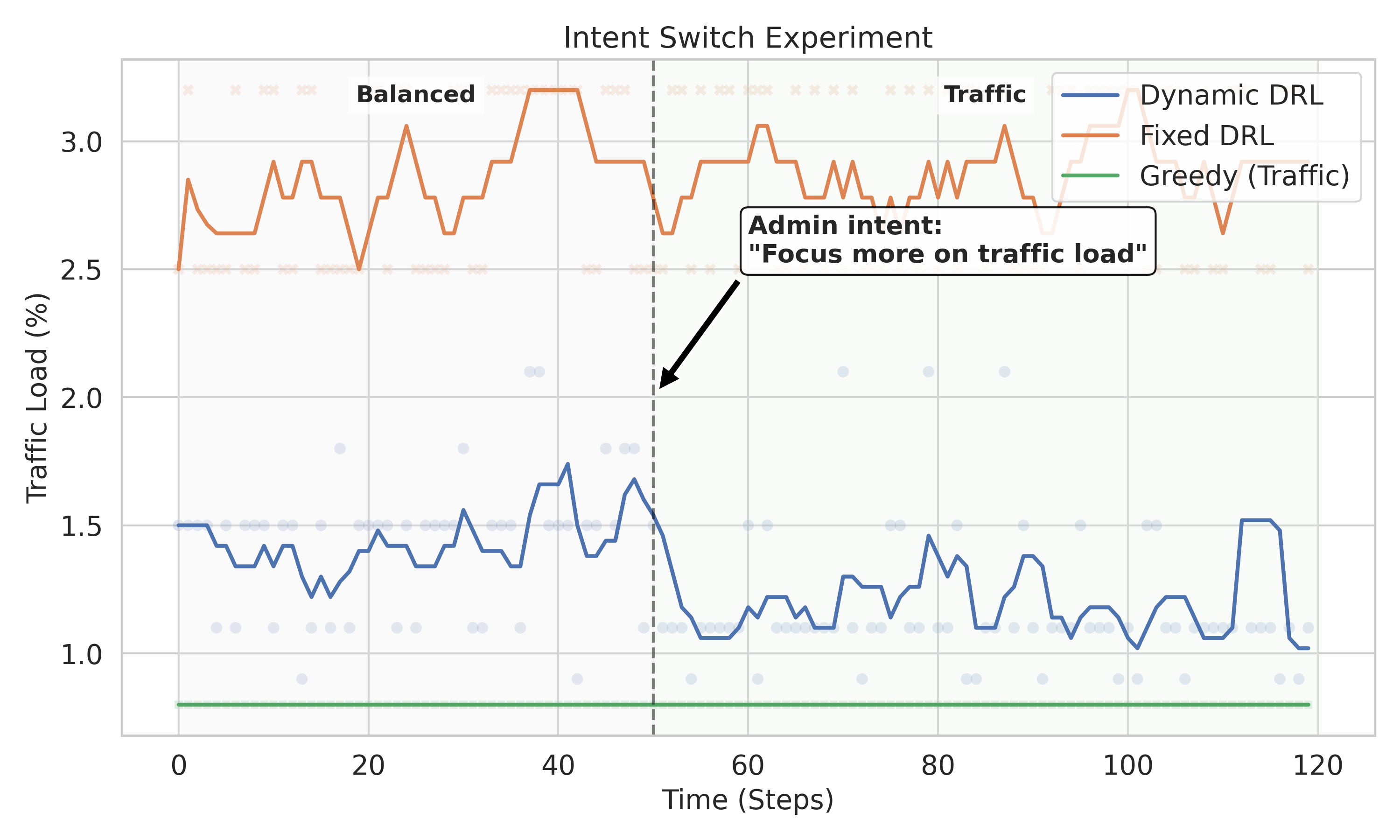}
    \caption{Intent adaptation from balanced to traffic-focused}
    \label{fig:intent}
\end{figure}

To demonstrate the adaptability of dynamic DRL, we evaluated all methods in a scenario where we changed the intent from balanced to traffic-focused by providing the new intent to the LLM interface. As illustrated in Figure \ref{fig:intent}, the LLM effectively mapped the high-level intent to the weight vector (0.15, 0.15, 0.70), and only the dynamic DRL scheme responded to the intent change, further decreasing traffic load by prioritizing UPFs with low traffic utilization.

\section{Conclusions and Future Work}
\label{sec:conc}
In this paper, we propose a multi-criteria DRL UPF selection method and validate it in a real-world experimental setup. Additionally, the proposed method supports adaptive intent and alerts translation into dynamic UPF selection using LLMs and NWDAF analytics. The results of the experiments demonstrate that our dynamic DRL solution outperforms the heuristic baselines and fixed DRL by balancing different metrics and considering multiple constraints. 

This work is limited to synthetic training metrics that do not fully capture mobility, failures, or large-scale conditions. Further evaluation is needed with stronger baselines and to assess robustness against malformed or adversarial LLM outputs. Future work will evaluate longer and more diverse workloads, larger UPF pools, mobility-aware scenarios, and deployment-scale LLM latency.


\bibliographystyle{IEEEtran}
\bibliography{biblio}

\begin{thebibliography}{10}
\providecommand{\url}[1]{#1}
\csname url@samestyle\endcsname
\providecommand{\newblock}{\relax}
\providecommand{\bibinfo}[2]{#2}
\providecommand{\BIBentrySTDinterwordspacing}{\spaceskip=0pt\relax}
\providecommand{\BIBentryALTinterwordstretchfactor}{4}
\providecommand{\BIBentryALTinterwordspacing}{\spaceskip=\fontdimen2\font plus
\BIBentryALTinterwordstretchfactor\fontdimen3\font minus \fontdimen4\font\relax}
\providecommand{\BIBforeignlanguage}[2]{{%
\expandafter\ifx\csname l@#1\endcsname\relax
\typeout{** WARNING: IEEEtran.bst: No hyphenation pattern has been}%
\typeout{** loaded for the language `#1'. Using the pattern for}%
\typeout{** the default language instead.}%
\else
\language=\csname l@#1\endcsname
\fi
#2}}
\providecommand{\BIBdecl}{\relax}
\BIBdecl

\bibitem{doorgakant2025end}
B.~Doorgakant, T.~P. Fowdur, and M.~O. Akinsolu, ``End-to-end power models for 5g radio access network architectures with a perspective on 6g,'' \emph{Mathematics}, vol.~13, no.~3, pp. 466--466, 2025.

\bibitem{husain20223gpp}
S.~Husain, A.~Kunz \emph{et~al.}, ``3gpp 5g core network: An overview and future directions.'' \emph{Journal of Information \& Communication Convergence Engineering}, vol.~20, no.~1, 2022.

\bibitem{alevizaki2021dynamic}
V.-M. Alevizaki, M.~Anastasopoulos, A.~Tzanakaki, D.~Simeonidou, and U.~Bristiol, ``Dynamic selection of user plane function in 5g environments,'' in \emph{Proc. of Optical Network Design and Modelling}, 2021.

\bibitem{leyva2020dynamic}
I.~Leyva-Pupo, C.~Cervell{\'o}-Pastor, C.~Anagnostopoulos, and D.~P. Pezaros, ``Dynamic scheduling and optimal reconfiguration of upf placement in 5g networks,'' in \emph{Proceedings of the 23rd International ACM Conference on Modeling, Analysis and Simulation of Wireless and Mobile Systems}, 2020, pp. 103--111.

\bibitem{yokobori2019user}
A.~Yokobori~S{\"a}v{\"o}, ``User plane selection for core networks using deep reinforcement learning,'' 2019.

\bibitem{bellin2024power}
A.~Bellin, N.~Di~Cicco, D.~Munaretto, and F.~Granelli, ``Power consumption-aware 5g edge upf selection using deep reinforcement learning,'' in \emph{2024 IEEE Conference on Network Function Virtualization and Software Defined Networks (NFV-SDN)}.\hskip 1em plus 0.5em minus 0.4em\relax IEEE, 2024, pp. 1--6.

\bibitem{barrachina2024demand}
S.~Barrachina-Munoz, R.~Nikbakht, A.~Bel, J.~Baranda, M.~Payar{\'o}, and J.~Mangues-Bafalluy, ``On-demand 5g/6g edge verticals via third-party upf selection and cloud-native relocation,'' in \emph{2024 IEEE Symposium on Computers and Communications (ISCC)}.\hskip 1em plus 0.5em minus 0.4em\relax IEEE, 2024, pp. 1--3.

\bibitem{3gpp}
{3GPP}, ``{Procedures for the 5G System (5GS)},'' {3rd Generation Partnership Project (3GPP)}, {Technical Specification (TS)} {23.502}, {July} {2021}, {Release 15}.

\bibitem{goshi2024joint}
E.~Goshi, H.~Harkous, S.~Ahvar, R.~Pries, F.~Mehmeti, and W.~Kellerer, ``Joint upf and edge applications placement and routing in 5g \& beyond,'' in \emph{2024 IEEE 21st International Conference on Mobile Ad-Hoc and Smart Systems (MASS)}.\hskip 1em plus 0.5em minus 0.4em\relax IEEE, 2024, pp. 446--452.

\bibitem{schulman2017proximal}
J.~Schulman, F.~Wolski, P.~Dhariwal, A.~Radford, and O.~Klimov, ``Proximal policy optimization algorithms,'' \emph{arXiv preprint arXiv:1707.06347}, 2017.

\bibitem{ueransim_2025}
A.~Güngör, ``{UERANSIM}: Open source 5g {UE} and {RAN} (gnodeb) simulator,'' \url{https://github.com/aligungr/UERANSIM}, 2025, accessed: 2025-09-27.

\bibitem{open5gs2025}
{Open5GS Contributors}, ``{Open5GS: Open Source 5G and EPC},'' \url{https://github.com/open5gs/open5gs}, 2025, accessed: 2025-09-26.

\end{thebibliography}
\end{document}